%% file: BC3E.tex
\newcommand{\vect} [1] {\boldsymbol{#1}}
\providecommand{\numberTblEq}[1]{\refstepcounter{tblEqCounter}\label{#1}\thetag{\thetblEqCounter}}
\newcounter{tblEqCounter} 

\documentclass[twoside,leqno,twocolumn]{article}
\usepackage{ltexpprt}
\usepackage{amsmath,amsfonts,amscd}
\usepackage{epsfig}         	
\usepackage{graphicx}
\usepackage{rotating}
\usepackage{tikz}
\usetikzlibrary{calc,through,backgrounds}
\usetikzlibrary{arrows,automata,positioning,fit}
\usetikzlibrary{decorations.pathreplacing}

\usepackage{algorithm}
\usepackage{algorithmic}
\usepackage{multirow}
\usepackage{hyperref}
\usepackage{url}

\begin{document}

\title{\Large Probabilistic Combination of Classifier and Cluster Ensembles for Non-transductive Learning}
\author{Ayan Acharya\thanks{University of Texas at Austin, Austin, TX, USA. Email: \{aacharya@, ghosh@ece\}.utexas.edu}\\
\and
Eduardo R. Hruschka\thanks{University of Sao Paulo at Sao Carlos, Brazil. Email: erh@icmc.usp.br}
\and
Joydeep Ghosh$^{\ast}$
\and
Badrul Sarwar\thanks{eBay Research Lab, San Jose, CA, USA. Email: \{bsarwar, jruvini\}@ebay.com}
\and
Jean-David Ruvini$^{\ddagger}$
}
\date{}

\maketitle


\begin{abstract} \small\baselineskip=9pt
Unsupervised models can provide supplementary soft constraints to help classify new target data under the assumption that similar objects in the target set 
are more likely to share the same class label. Such models can also help detect possible differences between training and target distributions, 
which is useful in applications where concept drift may take place. 
This paper describes a Bayesian framework that takes as input class labels from existing classifiers (designed based 
on labeled data from the source domain), as well as cluster labels 
from a cluster ensemble operating solely on the target data to be classified, and yields a consensus labeling of the target data.
This framework is 
particularly useful when the statistics of the target data drift or change from those of the training data. We 
also show that the proposed framework is privacy-aware and allows performing 
distributed learning 
when data/models have sharing restrictions.
Experiments show that our framework can yield superior results to those provided by applying classifier ensembles only.

\end{abstract}

\section{Introduction}

In several data mining applications, 
one builds an initial classification model that needs to be applied to unlabeled data acquired 
subsequently. Since the statistics of the underlying phenomena being modeled changes with time, these classifiers may also need to be occasionally 
rebuilt if performance degrades beyond an acceptable level. In such situations, it is desirable that the classifier functions well with as little 
labeling of new data as possible, since labeling can be expensive in terms of time and money, and a potentially error-prone process. Moreover, the 
classifier should be able to adapt to changing statistics to some extent, given the aforementioned constraints. 

This paper addresses the problem of combining multiple classifiers and clusterers in a fairly general setting, that includes the scenario sketched above.  
An  ensemble of classifiers is first learnt on an initial labeled training dataset after which the training data can be discarded.  
Subsequently, when new unlabeled target data is encountered, a cluster ensemble is applied to it, thereby generating cluster 
labels for the target data. 
The heart of our approach is a Bayesian framework that combines both sources of information (class/cluster labels) to yield a consensus labeling of the target data. 

The setting described above is, in principle, different from transductive learning setups where both labeled and unlabeled data are available at the same time for 
model building~\cite{sibe08}, as well as online methods \cite{blum98}.
Additional differences from existing approaches are described in the section on related works. For the moment we note that the underlying assumption is that similar new objects in the target set are more likely to share the same class label. Thus, the supplementary constraints provided by 
the cluster ensemble can be useful for improving the generalization capability of the resulting classifier system.
Also, these supplementary constraints 
can be useful for designing learning methods 
that help determining differences between training and target distributions, making the overall system more robust against concept drift. 

We also show that our approach can combine cluster and classifier ensembles in a privacy-preserving 
setting. 
This approach can be useful in a variety of applications. For example, the data sites can represent 
parties that are a group of banks, with their own sets of customers, who would like 
to have a better insight into the  behavior of the entire customer population without 
compromising the privacy of their individual customers.

The remainder of the paper is organized as follows. The next section addresses related work. 
The proposed Bayesian framework --- named \textbf{BC\textsuperscript{3}E}, from \textbf{B}ayesian \textbf{C}ombination 
of \textbf{C}lassifiers and \textbf{C}lusterer \textbf{E}nsembles --- is described in Section \ref{sec:BC3E}. 
Issues with privacy preservation are discussed in Section \ref{sec:privpres} and the experimental results are reported 
in Section \ref{exp}. Finally, Section \ref{conclusion} concludes the paper.

\section{Related Work}

The combination of multiple classifiers to generate an ensemble has been proven to be more useful compared to the use of individual classifiers \cite{oztu08}.
Analogously, several research efforts have shown that cluster ensembles can improve the quality of results as compared to a single clusterer --- {\it e.g.}, see \cite{wasb11} and references therein. 
Most of the motivations for combining ensembles of classifiers and clusterers are similar to those that hold for the standalone use of 
either classifier or cluster ensembles. 
Additionally, unsupervised 
models can provide supplementary constraints for classifying new data and thereby improve the generalization
capability of the resulting classifier. 
These successes provide the motivation for designing effective ways of leveraging 
both classifier and cluster ensembles to solve challenging prediction problems.

Specific mechanisms for combining classification and clustering models however 
have been introduced only recently in the Bipartite Graph-based Consensus 
Maximization (\textbf{BGCM}) algorithm \cite{galf09}, the Locally Weighted Ensemble (\textbf{LWE}) algorithm \cite{gafj08} 
and, in the \textbf{C\textsuperscript{3}E}  algorithm \cite{achg11}. 
Both \textbf{BGCM} and \textbf{C\textsuperscript{3}E} have parameters that control the relative importance of classifiers and clusterers.
In traditional semi-supervised settings, such parameters can be optimized via cross-validation. 
However, if the training and the target distributions are different, cross-validation is not possible.
From this viewpoint, our approach (\textbf{BC\textsuperscript{3}E}) can be seen as an extension of \textbf{C\textsuperscript{3}E} \cite{achg11} that 
is capable of dealing with this issue in a more principled way. In addition, the algorithms in \cite{galf09,gafj08,achg11} do not deal with privacy issues, 
whereas our probabilistic framework can combine class labels with cluster labels under conditions where sharing of individual records across data sites 
is not permitted. It uses a soft probabilistic notion of privacy, based on a quantifiable information-theoretic formulation 
\cite{megh03}. Note that existing works on Bayesian classifier ensembles --- \textit{e.g.}, \cite{edvi09,ChGeMc06,GhHy03} --- 
do not deal with privacy issues. 

From the clustering side, the proposed model borrows ideas from the Bayesian 
Cluster Ensemble \cite{wasb11}.
In \cite{achg11a}, we introduced some preliminary ideas that are further developed in our current paper. In particular, 
the algorithm in \cite{achg11a} is not capable of automatically estimating the importance that classifiers and clusterers should have. This 
property is fundamental for applications where training and target distributions are different. In addition, the Bayesian 
model presented here is considerably different and requires more sophisticated inference and estimation procedures.

\section{Probabilistic Model}
\label{sec:BC3E}

We assume that a classifier ensemble has been (previously) induced from a training set.
At this point and assuming a non-transductive setting, the training data can be discarded if so desired. Such a classifier ensemble
is employed to generate a number of class labels (one from each
classifier) for every object in the target set.
\textbf{BC\textsuperscript{3}E} refines such classifier prediction with the help of a cluster ensemble.
Each base clustering algorithm that is part of the
ensemble partitions the target set, providing cluster labels for each of its objects. From this
point of view, the cluster ensemble provides supplementary constraints for classifying
those objects, with the rationale that similar objects --- those that are likely to be clustered together
across (most of) the partitions that form the cluster ensemble --- are more likely to share the same class label.

\begin{figure*}[ht]
\begin{minipage}[b]{0.55\linewidth}
\centering
  \includegraphics[bb = 0 0 685 392, scale=0.35]{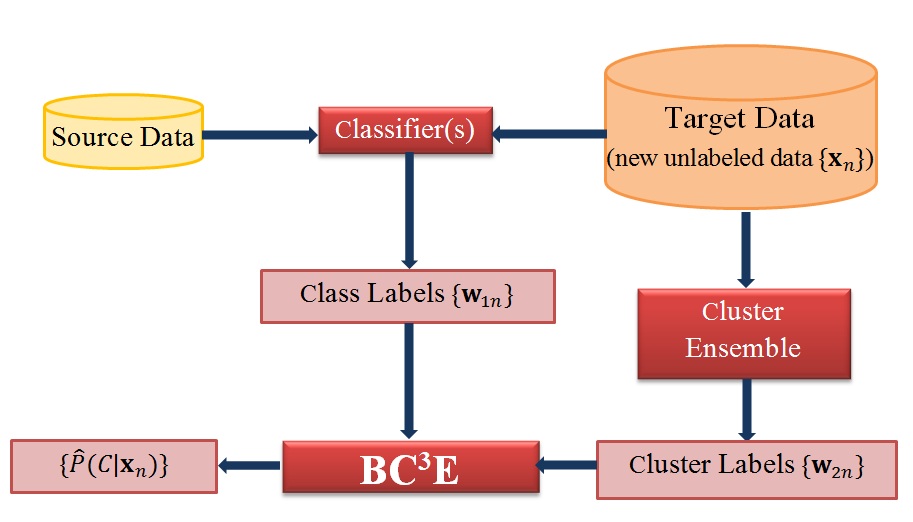}
  \caption{Combining Classifiers and Clusterers.}
 \label{fig:BC3E}
\end{minipage}
\hspace{0.5cm}
\begin{minipage}[b]{0.45\linewidth}
\centering
\scalebox{0.73}{\input{BC3E_fig.tex}}
\caption{Graphical Model for \textbf{BC\textsuperscript{3}E}}
\label{figure_BC3E_GM}
\end{minipage}
\end{figure*}

Consider a target set $\mathcal{X}=\{\mathbf{x}_{n}\}_{n=1}^{N}$ formed by $N$ unlabeled objects. A classifier
ensemble composed of $r_{1}$ models has produced $r_{1}$ class labels
for every
object $\mathbf{x}_{n}\in\mathcal{X}$. It is assumed that the target objects belong to $k$ classes denoted
by $C=\{C_{i}\}_{i=1}^{k}$ and at least one object from each of these classes was observed in the training 
phase (\emph{i.e.} we do not consider ``novel'' classes in the target set). Similarly, consider that a cluster ensemble comprised of $r_{2}$ 
clustering algorithms has generated cluster labels for every object in the target set.
The number of clusters need not be the same across different clustering algorithms.
Also, it should be noted that the cluster labeled as \textit{1} in a given data partition may not
align with the cluster numbered \textit{1} in another partition,
and none of these clusters may correspond to class \textit{1}. Given the class and cluster labels, the objective is to come up with
refined class probability distributions $\{(\hat{P}(C_{i}|\mathbf{x}_{n}))_{i=1}^{k} = \vect{y}_{n}\}_{n=1}^{N}$ of the target
set objects. This framework is illustrated in Fig. \ref{fig:BC3E}.

The observed class and cluster labels are represented as $\vect{W}=\{\{\vect{w}_{1nl}\},\{\vect{w}_{2nm}\}\}$ where $\vect{w}_{1nl}$
is the 1-of-$k$ representation of class label of the $n^{\text{th}}$ object given by the $l^{\text{th}}$ classifier,
and $\vect{w}_{2nm}$ is the 1-of-$k^{(m)}$ representation of cluster label assigned to the $n^{\text{th}}$ object by the $m^{\text{th}}$ clusterer.
A generative model is proposed to explain the observations $\vect{W}$, where each object $\mathbf{x}_{n}$
has an underlying mixed-membership to the $k$ different classes. Let $f(\vect{y}_{n})$ denote the latent
mixed-membership vector for $\mathbf{x}_{n}$, where $f(\mathbf{x})= \frac{\text{exp}(x_{i})}{\sum_{i=1}\text{exp}(x_{i})}$ is the softmax function.
$\vect{y}_{n}$ is sampled from a normal distribution $\mathcal{N}(\vect{\mu}, \vect{\Sigma})$.
Also, corresponding to the $i^{\text{th}}$ class and $m^{\text{th}}$ base clustering,
we assume a multinomial distribution $\vect{\beta}_{mi}$ over the cluster labels of the $m^{\text{th}}$ base clustering.
Therefore, $\vect{\beta}_{mi}$ is of dimension $k^{(m)}$ and $\sum_{j=1}^{k^{(m)}}\vect{\beta}_{mij}=1$ if the $m^{\text{th}}$ base
clustering has $k^{(m)}$ clusters. The data generative process, whose corresponding graphical model is shown in \ref{figure_BC3E_GM}, can be summarized as follows.

For each $\mathbf{x}_n\in\mathcal{X}$:
\begin{enumerate}
 \item Choose $\vect{y}_{n}\sim\mathcal{N}(\vect{\mu}, \vect{\Sigma})$, where $\vect{\mu}\in\mathbb{R}^{k}$ is the mean 
 and $\vect{\Sigma}\in\mathbb{R}^{k\times k}$ is the covariance.
 \item Choose $\vect{\theta}_{n}\sim\mathcal{N}(\vect{y}_{n}, \delta^{2}I_{k})$, where $\delta^{2}\ge 0$ is the scaling 
 factor of the covariance of the normal distribution centered at $\vect{y}_{n}$, and $I_{k}$ is the identity $k \times k$ matrix.
 \item $\forall l\in\{1,2,\cdots,r_{1}\}$, choose $\vect{w}_{1nl}\sim f(\vect{y}_{n})$.
 \item $\forall m\in\{1,2,\cdots,r_{2}\}$:
 \begin{enumerate}
 \item Choose $\vect{z}_{nm}\sim f(\vect{\theta}_{n})$, where $\vect{z}_{nm}$ is a $k$-dimensional vector with 1-of-$k$ representation.
 \item Choose $\vect{w}_{2nm}\sim \text{multinomial}(\vect{\beta}_{r\vect{z}_{nm}})$.
 \end{enumerate}
\end{enumerate}

The observed class labels $\{\vect{w}_{1nl}\}$ are assumed to be sampled from the latent mixed-membership vector $f(\vect{y}_{n})$.
If the $n^{\text{th}}$ object is sampled from the $i^{\text{th}}$ class in the $m^{\text{th}}$ base clustering (implying $z_{nmi}=1$),
then its cluster label will be sampled from the multinomial distribution $\vect{\beta}_{mi}$. This particular generative process is analogous
to the one used by the Bayesian Cluster Ensemble in \cite{wasb11}. The fact that $\vect{\theta}_{n}$ is sampled from
$\mathcal{N}(\vect{y}_{n}, \delta^{2}I_{k})$ needs further
clarification. In practice, the observed class labels and cluster labels carry different intrinsic weights.
If the observations from the classifiers are assigned too much weight compared to those from clustering, there is little hope for the clustering to enhance classification.
Similarly, if the observations from the clustering are given too much of importance, the classification performance might deteriorate.
Ideally, the unsupervised information is only expected
to enhance the classification accuracy.

Aimed at building a ``safe'' model that can intelligently utilize or reject the unsupervised information, $\vect{\theta}_{n}$ is sampled from $\mathcal{N}(\vect{y}_{n}, \delta^{2}I_{k})$ where the parameter $\delta$ decides how much the observations from the clusterings can be trusted.
If $\delta^{2}$ is a large positive number, $\vect{y}_{n}$ does not
have to explain the posterior of $\vect{\theta}_{n}$.
From the generative model perspective, this means that the sampled value
of $\vect{\theta}_{n}$ is not governed by $\vect{y}_{n}$ anymore as the distribution has very large variance.
On the other hand, if $\delta^{2}$ is a small positive number, $\vect{y}_{n}$ has to explain
the posterior of $\vect{\theta}_{n}$ and hence the observations from the clustering.
Therefore, the posteriors of
$\{\vect{y}_{n}\}$ are expected to get more accurate compared to the case if they only had to explain the classification results.
A concrete quantitative argument for this intuitive statement will be presented later.

To address the log-likelihood function of \textbf{BC\textsuperscript{3}E}, let us denote the set of hidden variables by
$\vect{Z}= \{\{\vect{y}_{n},\{\vect{\theta}_{n}\},\{\vect{z}_{nm}\}\}$. The model parameters can conveniently be represented
by $\vect{\zeta}_{0}=\{\vect{\mu},\vect{\Sigma},\delta^{2},\{\vect{\beta}_{mi}\}\}$.
The joint distribution of the hidden and observed variables can be written as:
\begin{equation}
\label{incmpllkhd}
 p(\mathbf{X},\vect{Z}|\vect{\zeta}_{0}) =
\prod_{n=1}^{N}p(\vect{y}_{n}|\vect{\mu}, \vect{\Sigma})p(\vect{\theta}_{n}|\vect{y}_{n}, \delta^{2}I_{k}).
\end{equation}
\begin{equation*}
\prod_{l=1}^{r_{1}}p(w_{1nl}|f(\vect{y}_{n}))\prod_{m=1}^{r_{2}}p(\vect{z}_{nm}|f(\vect{\theta}_{n}))
p(w_{2nm}|\vect{\beta}, \vect{z}_{nm})
\end{equation*}
The inference and estimation is performed using Variational Expectation-Maximization (\textbf{VEM}) to avoid
computational intractability due to the coupling between $\vect{\theta}$ and $\vect{\beta}$.

\subsection{Approximate Inference and Estimation:}
\label{sec:infnest}
%
\subsubsection{Inference:}
\label{inference}
To obtain a tractable lower bound on the observed log-likelihood, we specify a fully factorized distribution to approximate the true
posterior of the hidden variables:
\begin{eqnarray}
\label{factorized}
 q(\vect{Z}|\{\vect{\zeta}_{n}\}_{n=1}^{N})
&=&\prod_{n=1}^{N}
q(\vect{y}_{n}|\vect{\mu}_{n}, \vect{\Sigma}_{n})
q(\vect{\theta}_{n}|\vect{\epsilon}_{n}, \vect{\Delta}_{n})\nonumber\\
&& \prod_{m=1}^{r_{2}}q(\vect{z}_{nm}|\vect{\phi}_{nm})
\end{eqnarray}
where  $\vect{y}_{n}\sim\mathcal{N}(\vect{\mu}_{n}, \vect{\Sigma}_{n})$,
$\vect{\theta}_{n}\sim\mathcal{N}(\vect{\epsilon}_{n}, \vect{\Delta}_{n})\text{ }\forall n\in\{1,2,\cdots, N\}$,
$\vect{z}_{nm}\sim \text{multinomial}(\vect{\phi}_{nm})
\text{ }\forall n\in\{1,2,\cdots, N\}$ and $\forall m\in\{1,2,\cdots,r_{2}\}$, and
$\vect{\zeta}_{n}=\{\vect{\mu}_{n}, \vect{\Sigma}_{n}, \vect{\epsilon}_{n}, \vect{\Delta}_{n}), \{\vect{\phi}_{nm}\}\}$ -- the
set of variational parameters corresponding to the $n^{\text{th}}$ object.
Further, $\vect{\mu}_{n}, \vect{\epsilon}_{n}\in\mathbb{R}^{k}$, $\vect{\Sigma}_{n}, \vect{\Delta}_{n}\in \mathbb{R}^{k \times k}$ $\forall n$
and $\vect{\phi}_{nm}=(\phi_{nmi})_{i=1}^{k}\text{ }\forall n,m$; where the components of
the corresponding vectors are made explicit. To work with 
less parameters, all the covariance matrices are assumed to be diagonal.
Therefore, $\vect{\Sigma} = \text{diag}\left((\sigma_{i})_{i=1}^{k}\right)$,
$\vect{\Sigma}_{n} = \text{diag}\left((\sigma_{ni})_{i=1}^{k}\right)$,
and $\vect{\Delta}_{n} = \text{diag}\left((\delta_{ni})_{i=1}^{k}\right)$.
Using Jensen's inequality, a lower bound on the observed log-likelihood can be derived as:
\begin{eqnarray}
\label{lwbound}
 \text{log} [p(\vect{X}|\vect{\zeta}_{0})] &\ge&
\mathbf{E}_{q(\vect{Z})}\left[\text{log} [p(\vect{X},\vect{Z}|\vect{\zeta}_{0})]\right]
+H(q(\vect{Z}))\nonumber\\
&=&\mathcal{L}(q(\vect{Z}))
\end{eqnarray}
where $H(q(\vect{Z}))=-\mathbf{E}_{q(\vect{Z})}[\text{log} [q(\vect{Z})]]$ is the entropy of the variational distribution $q(\vect{Z})$,
and $\mathbf{E}_{q(\vect{Z})}[.]$ is the expectation w.r.t $q(\vect{Z})$.

Let $\mathbf{\mathcal{Q}}$ be the set
of all distributions having a fully factorized form as given in (\ref{factorized}). The optimal
distribution that produces the tightest possible lower bound $\mathcal{L}$ is given by:
\begin{eqnarray}
\label{lwboundopt}
 q^{*}&=& \arg\min_{q\in\mathbf{\mathcal{Q}}}\text{KL}(p(\vect{Z}|\vect{X},\vect{\zeta}_{0})||q(\vect{Z})).
\end{eqnarray}

In equations (\ref{updatephi}), (\ref{updatekappa}), (\ref{updatexi}), (\ref{updatedeltan2}), (\ref{updatemun}), (\ref{updatesigman2}) 
and (\ref{updateepsilon}) in Table \ref{tab:updateeqns},
the optimal values of the variational parameters that satisfy (\ref{lwboundopt}) are presented.
Since the logistic normal distribution is not conjugate to multinomial, the update equations of all the parameters cannot be obtained in closed form.
For the parameters that do not have a closed form solution for the update, we just present the part of the objective function that depends on the concerned parameter and some numeric optimization method has to be used for optimizing the lower bound. Since $\vect{\phi}_{nm}$ is a multinomial distribution, the updated
values of the $k$ components should be normalized to unity.
Note that the optimal value of one of the variational parameters depends on the others and, therefore, an iterative optimization is adopted to minimize the 
lower bound till convergence is achieved.

\setcounter{tblEqCounter}{\theequation}
\begin{table*}[tbp]
\centering
\begin{tabular}{||c||c||}
\hline
\multicolumn{2}{||c||}{Update Equations}\\\hline
$\phi_{nmi}^{*}\propto \text{exp}\left(\epsilon_{ni} + \displaystyle\sum_{j=1}^{k^{(m)}}\beta_{mij}w_{2nmj}\right) \text{ }\forall n,m,i.$ \numberTblEq{updatephi}
& $\vect{\mu}^{*} =  \frac{1}{N}\displaystyle\sum_{n=1}^{N}\vect{\mu}_{n}$. \numberTblEq{updatemu} \\
\hline
$\kappa_{n}^{*} =  \displaystyle\sum_{i=1}^{k}\text{exp}(\mu_{ni} + \sigma_{ni}^2/2) \text{ }\forall n.$ \numberTblEq{updatekappa}
& $\beta_{mij}^{*}\propto\displaystyle\sum_{n=1}^{N}\phi_{nmi}w_{2nmj}\text{ }\forall j\in{1,2,\cdots,k^{m}}.$ \numberTblEq{updatebeta}\\
\hline
$ \xi_{n}^{*} =  \displaystyle\sum_{i=1}^{k}\text{exp}(\epsilon_{ni} + \delta_{ni}^2/2) \text{ }\forall n.$ \numberTblEq{updatexi}
& $\delta^{2} =  \frac{1}{Nk}\displaystyle\sum_{n=1}^{N}\displaystyle\sum_{i=1}^{k} \big[(\epsilon_{ni} - \mu_{ni})^{2} + \sigma_{ni}^{2} + \delta_{ni}^{2}\big].$
\numberTblEq{updatedelta2}\\
\hline
\scriptsize $\mathcal{L}_{[\vect{\delta}_{n}^{2}]} =  -\frac{1}{2}\displaystyle\sum_{i=1}^{k} \frac{\delta_{ni}^2}{\delta^2} - \frac{1}{2}\displaystyle\sum_{i=1}^{k} \text{log}(\delta_{ni}^2)
 - \frac{r_{2}}{\xi_{n}}\displaystyle\sum_{i=1}^{k}\text{exp}(\epsilon_{ni}+\delta_{ni}^{2}/2).$ \numberTblEq{updatedeltan2}
& \scriptsize $\mathcal{L}_{[\vect{\sigma}^{2}]} = -\frac{N}{2}\displaystyle\sum_{i=1}^{k} \text{log}(\sigma_{i}^{2}) - \frac{1}{2}\displaystyle\sum_{n=1}^{N}\displaystyle\sum_{i=1}^{k}
\big[\frac{\sigma_{ni}^2 + (\mu_{ni}-\mu_{i})^2}{\sigma_{i}^{2}}\big].$ \numberTblEq{updatesigma2}\\
\hline
\multicolumn{2}{||c||} {$\mathcal{L}_{[\vect{\mu}_{n}]} =  -\frac{1}{2}\displaystyle\sum_{i=1}^{k} \frac{(\mu_{ni}-\mu_{i})^2}{\sigma_{i}^2} - \frac{1}{2\delta^{2}}\displaystyle\sum_{i=1}^{k}(\mu_{ni}-\epsilon_{ni})^2
+ \displaystyle\sum_{l=1}^{r_{1}}\displaystyle\sum_{i=1}^{k}w_{1nli}\mu_{ni} - \frac{r_{1}}{\xi_{n}}\displaystyle\sum_{i=1}^{k}\text{exp}(\mu_{ni}+\sigma_{ni}^{2}/2).$ \numberTblEq{updatemun}}\\
\hline
\multicolumn{2}{||c||} {$\mathcal{L}_{[\vect{\sigma}_{n}^{2}]} =  -\frac{1}{2}\displaystyle\sum_{i=1}^{k} \frac{\sigma_{ni}^2}{\sigma_{i}^2} - \frac{1}{2}\displaystyle\sum_{i=1}^{k} \text{log}(\sigma_{ni}^2)
- \frac{1}{2}\displaystyle\sum_{i=1}^{k} \frac{\sigma_{ni}^2}{\delta^{2}} - \frac{r_{1}}{\kappa_{n}}\displaystyle\sum_{i=1}^{k}\text{exp}(\mu_{ni}+\sigma_{ni}^{2}/2).$ \numberTblEq{updatesigman2}}\\
\hline
\multicolumn{2}{||c||} {$\mathcal{L}_{[\vect{\epsilon}_{n}]} = \displaystyle\sum_{m=1}^{r_{2}}\displaystyle\sum_{i=1}^{k}\phi_{nmi}\epsilon_{ni} - \frac{1}{\xi_{n}}\displaystyle\sum_{i=1}^{k}\text{exp}(\epsilon_{ni}+\delta_{ni}^{2}/2)
- \frac{1}{2}\displaystyle\sum_{i=1}^{k}\frac{(\epsilon_{ni}-\mu_{ni})^2}{\delta^2}.$ \numberTblEq{updateepsilon}}\\
\hline
\end{tabular}
\caption{Equations for update of variational and model parameters in \textbf{BC\textsuperscript{3}E}}
\label{tab:updateeqns}
\end{table*}
\setcounter{equation}{\thetblEqCounter}

Equations (\ref{updatekappa}) and (\ref{updatexi}) present updates for two new parameters. These parameters come from $\mathbb{E}_{q}(\text{log }p(\vect{w}_{1nl}|f(\vect{y}_{n})))$ and $\mathbb{E}_{q}(\text{log }p(\vect{z}_{nm}|f(\vect{\theta}_{n})))$ respectively.
Both of these integrations do not have analytic solution and hence a first order Taylor approximation is utilized as also done in
\cite{blla07}.
A closer inspection of (\ref{updatemun}) reveals that $\delta^{2}$ appears in the denominator of the term
$\displaystyle\sum_{i=1}^{k}(\mu_{ni}-\epsilon_{ni})^2/\delta^{2}$ in the objective.
Hence, larger values of $\delta^2$ will nullify any effect from $\vect{\epsilon}_{n}$ which, in turn, is affected by the observations $\{\vect{w}_{2nm}\}$
(as is obvious from (\ref{updateepsilon})).
On the other hand, if $\delta^2$ is small enough, $\vect{\epsilon}_{n}$ can strongly impact the values of $\vect{\mu}_{n}$.

\subsubsection{Estimation:}

For estimation, we maximize the optimized lower bound obtained
from the variational inference w.r.t the free
model parameters $\vect{\zeta}_{0}$ (by keeping the variational parameters fixed).
The optimal values of the model parameters are presented in equations (\ref{updatemu}), (\ref{updatebeta}) and (\ref{updatedelta2}).
Since $\vect{\beta}_{mi}$ is a multinomial distribution, the updated values of $k^{(m)}$ components should be normalized to unity.
However, no closed form of update exists for $\vect{\sigma}^2$, and a numeric optimization method has to be resorted to. The part of the objective function that
depends on $\vect{\sigma}^2$ is provided in Eq. (\ref{updatesigma2}).
Once the optimization in M-step is done, E-step starts and the iterative update is continued till convergence.
The variational parameters $\{\vect{\mu}_{n}\}_{n=1}^{N}$ are then investigated which
serve as proxy for the refined posterior estimates of $\{\vect{y}_{n}\}_{n=1}^{N}$. The main steps of inference and estimation are
concisely presented in Algorithm \ref{algo:LearnBC3E}.
\input{algorithm_BC3E.tex}

\section{Privacy Preserving Learning}
\label{sec:privpres}


Most of the privacy-aware distributed data mining techniques
 developed so far have focused on classification
or on  association rules \cite{agag01,evsr02}. 
There has also been some work on distributed clustering for {\em vertically partitioned data }
(different sites contain different attributes/features of a common set of records/objects) 
\cite{joka99}, and on parallelizing clustering  algorithms for {\em
horizontally partitioned data} (i.e. the objects are distributed amongst
the sites, which record the same set of features for
each object) \cite{dhmo99}. These techniques, however, do not specifically address 
privacy issues, other than through encryption \cite{vacl03}.

This is also true of earlier, data-parallel methods \cite{dhmo99} that are susceptible to privacy 
breaches, and also need a central planner that dictates what algorithm runs on each site. 
Finally, recent works on distributed differential privacy focus on query processing rather than data mining \cite{chrf12}.

In the sequel, we show that  the inference and estimation in 
\textbf{BC\textsuperscript{3}E} using \textbf{VEM} allows solving 
the cluster ensemble problem in a way that preserves privacy. Depending on how the objects with their cluster/class 
labels are distributed in different ``data sites'', we can have three scenarios -- i) Row Distributed Ensemble,
ii) Column Distributed Ensemble, and iii) Arbitrarily Distributed Ensemble. 
  
\subsection{Row Distributed Ensemble:}

In the row distributed ensemble learning framework, the test set $\mathcal{X}$ is partitioned into $D$ parts
and different parts are assumed to be at different locations. The objects from partition $d$ are denoted by
$\mathcal{X}_{d}$ so that $\mathcal{X}=\cup_{d=1}^{D}\mathcal{X}_{d}$. 
Now, a careful look at the E-step equations reveal that the update of variational parameters corresponding 
to each object in a given iteration is independent of those
of other objects. 
Therefore, we can maintain a client-server based framework where the
server only updates the model parameters (in the M-step) and the clients (there should be as many number of clients as there are 
distributed data sites) update the variational parameters.

For instance, consider a situation where a dataset is partitioned into two subsets $\mathcal{X}_{1}$ and $\mathcal{X}_{2}$
and these two subsets are located in two different data sites. Data site $1$ has access to $\mathcal{X}_{1}$
and a set of clustering and classification results pertaining to objects belonging to $\mathcal{X}_{1}$.
Similarly, data site $2$ has access to $\mathcal{X}_{2}$
and a set of clustering and classification results corresponding to $\mathcal{X}_{2}$. Further assume that
a set of distributed classification (clustering) algorithms were used to generate the class (cluster) labels
of the objects belonging to each set. Now, data site $1$ can update the variational parameters
$\vect{\zeta}_{n}$, $\forall \mathbf{x}_{n}\in \mathcal{X}_{1}$. Similarly, 
data site $2$ can update the variational parameters for all objects 
$\mathbf{x}_{n}\in \mathcal{X}_{2}$. Once the variational parameters are updated
in the E-step, the server gathers information from two sites and updates the model parameters. 
Now, a closer
inspection of the M-step update equations reveals that 
each of them contains a summation over the objects. Therefore,
individual data sites can send only some collective information to the server without transgressing privacy. 
For example, consider the update equation for $\beta_{mij}$. 
Eq. (\ref{updatebeta}) can be broken as follows:
\begin{equation}
\label{decomposition}
{\beta_{mij}}^{*}\propto\displaystyle\sum_{x_{n}\in\mathcal{X}_{1}}\phi_{nli}w_{2nli} + \displaystyle\sum_{x_{n}\in\mathcal{X}_{2}}\phi_{nli}w_{2nli}
\end{equation}
The first and second terms can be calculated in data sites $1$ and $2$ separately and sent to the server where the two 
terms can be added and $\beta_{mij}$ can get updated $\forall m,i,j$. 
Similarly, the other M-step update equations (performed by the server in an analogous way) 
also do not reveal any information about class or cluster labels of objects belonging to different data sites.
 
\subsection{Column Distributed Ensemble:}

In the column distributed framework, different data sites share the same set of objects 
but only a subset of base clusterings or 
classification results are available to each data site. 
For example, consider that we have two data sites and four sets of class and cluster labels and each 
data site has access to only two sets of classification
or clustering results. 
Assume that data site $1$ has access to the $1^\text{st}$ and $2^\text{nd}$ classification and clustering results and data site $2$ has access to the rest of the results.
As in the earlier case, a single server and two clients (corresponding to two different data sites) are maintained. Since each data site has access 
to all the objects, it is necessary to share the variational parameters corresponding to these objects. Therefore, 
$\{\kappa_{n}, \xi_{n}, \vect{\mu}_{n}, \vect{\sigma}_{n}, \vect{\epsilon}_{n}, \vect{\delta}_{n}\}_{n=1}^{N}$ are all updated in the server 
(which is accessible from each client). 

The site (and object) specific variational parameters $\{\phi_{nmi}\}$, however, cannot be shared and should be updated in individual sites.
This means that the updates (\ref{updatekappa}), (\ref{updatexi}), (\ref{updatemun}), (\ref{updateepsilon}),
(\ref{updatedeltan2}) and (\ref{updatesigman2}) should be performed in the server. 
On the other hand, the update for $\{\phi_{nmi}\} \forall n,i\text{ and } m\in\{1,2\}$ (corresponding to the 
$1^{\text{nd}}$ and $2^{\text{nd}}$ clustering or classification results) should be performed in data site $1$. 
Similarly, the update for $\{\phi_{nmi}\}$ $\forall n,i \text{ and } m\in\{3,4\}$ has to be performed in data site $2$. 
However, while updating $\{\vect{\mu}_{n}\}$, the calculation of the term
$\displaystyle\sum_{l=1}^{r_{1}}\displaystyle\sum_{i=1}^{k}w_{1nli}\mu_{ni}$ has to be performed without revealing the class labels
$\{\vect{w}_{1nl}\}$ to the server. To that end, it can be rewritten as:
\begin{equation}
 \displaystyle\sum_{l=1}^{r_{1}}\displaystyle\sum_{i=1}^{k}w_{1nli}\mu_{ni}=
 \displaystyle\sum_{l=1}^{2}\displaystyle\sum_{i=1}^{k}w_{1nli}\mu_{ni} + 
 \displaystyle\sum_{l=3}^{4}\displaystyle\sum_{i=1}^{k}w_{1nli}\mu_{ni},
\end{equation}
where the first term can be computed in data site $1$ and the second term can be computed by data site $2$ and then
can be added in the server. It can be seen that $\{\vect{w}_{1nl}\}$ can never be recovered by the server and 
hence privacy is ensured in the updates of the E-step.
Except for $\{\beta_{mij}\}$, all other model
parameters can be updated in the server in the M-step. However, the parameters $\{\beta_{mij}\}$ have to be updated separately inside the clients. 
Since $\{\beta_{mij}\}$ do not appear in any update equation performed in the server, there is no need to send these parameters to the server either. 
Therefore, in essence, the clients update the parameters $\{\phi_{nmi}\}$ and $\{\beta_{mij}\}$ in E-step and M-step respectively, and the server 
updates the remaining parameters. 

\subsection{Arbitrarily Distributed Ensemble:}

 In an arbitrarily distributed ensemble, each data site has access to only a subset of the data points or a subset of the
classification and clustering results. Fig. \ref{fig:arbpart} shows a situation with arbitrarily distributed
ensemble with six data sites. 

We now refer to Fig. \ref{fig:case4Fig} and explain the privacy preserved EM update for this setting. As before, corresponding
to each different data site, a client node is created. Clients that share a subset of the objects should have access to the
variational parameters corresponding to common objects. To highlight the sharing of objects by clients, the
test set $\mathcal{X}$ is partitioned into four subsets --- $\mathcal{X}_{1},\mathcal{X}_{2},\mathcal{X}_{3}\text{ and}\mathcal{X}_{4}$ as shown in 
Fig. \ref{fig:arbpart}. Similarly, the columns are also partitioned into three subsets: $G_{1}, G_{2}$, and $G_{3}$. 

Now, corresponding to each row partition, an ``Auxiliary Server''(AS) node is created. Each AS updates the variational parameters corresponding
to a set of shared objects. For example, in Fig. \ref{case4Fig}, AS$_1$ updates the variational parameters corresponding to $\mathcal{X}_{1}$
(using equations (\ref{updatexi}), (\ref{updatekappa}), (\ref{updatemun}), (\ref{updatesigman2}), (\ref{updateepsilon}), and (\ref{updatedeltan2})).
However, any variational parameter that is specific to both an object and a column is updated separately inside the corresponding
client (and hence it is connected with $C_{1}$ and $C_{2}$). Therefore, $\{\phi_{nmi}:n\in\mathcal{X}_{1}, m\in G_{1}\}$ are updated inside client 1 and 
$\{\phi_{nmi}:n\in\mathcal{X}_{1}, m\in G_{2}\cup G_{3}\}$ are updated inside client 2 (using Eq. (\ref{updatephi})). 
Once all variational parameters are updated in the E-step, M-step starts. Corresponding to each column partition,
an ``Auxiliary Client'' (AC) node is created. This node updates the model parameters $\beta_{mij}$ (using Eq. (\ref{updatebeta})) which are 
specific to columns belonging to $G_{1}$. Since $C_{1}$, $C_{3}$, and $C_{5}$ share the columns from the subset $G_{1}$, $AC_{1}$ is connected with these three nodes
in Fig. \ref{case4Fig}. The remaining model parameters are, however, updated in a ``Server'' (using equations (\ref{updatemu}), (\ref{updatedelta2}), (\ref{updatesigma2})).

\begin{figure*}[ht]
\begin{minipage}[b]{0.30\linewidth}
\scalebox{0.8}{\input{figarbd.tex}}
\caption{Arbitrarily Distributed Ensemble}
\label{fig:arbpart}
\end{minipage}
\hspace{0.5cm}
\begin{minipage}[b]{0.65\linewidth}
\centering
 \scalebox{0.8}{\input{case4fig.tex}}
 \caption{Parameter Update for Arbitrarily Distributed Ensemble} 
 \label{fig:case4Fig}
\end{minipage}
\end{figure*}

In Fig. \ref{case4Fig}, the bidirectional edges indicate that messages are sent
to and from the connecting nodes. We have avoided separate arrows for each direction only to keep the 
figure uncluttered.
The edges are also numbered near to their origin. For a comprehensive understanding of the privacy
preservation, the messages transfered through each edge have also been enlisted in the supplementary material. 
The messages sent from the auxiliary servers
to the main server are of the form given in Eq. (\ref{decomposition}) and are denoted as ``partial sums''.
Expectedly, messages sent out from
a client node are ``masked'' in such a way that no other node can decode the cluster labels or class labels of points belonging to
that client. 
This approach is completely general and will work for any arbitrarily 
partitioned ensemble given that each partition contains at least two sets of classification results. 
Note that the ACs and ASs are only helpful
in conceptual understanding of the parameter update and sharing. In practice, there is no 
real need for these extra storage devices/locations. Client nodes can themselves take the place of ASs and ACs and 
even the main server as long as the updates are performed in proper sequence\footnote{Note that such framework allows running 
the updates of the same stage in parallel in different sites, thereby saving the computation time in large scale implementations.}.

\section{Experiments}
\label{exp}
In this section, two different sets of experiments are reported. The first set 
is for transfer learning with a text classification data from eBay Inc.
The other set is for non-transductive semisupervised learning where some 
publicly available datasets are used to simulate the working 
environment of \textbf{BC\textsuperscript{3}E}.

\subsection{Transfer Learning:}
To show the capability of \textbf{BC\textsuperscript{3}E} in solving transfer learning problems, 
we use a large scale text classification dataset from eBay Inc.
The training data consists of 83 million items sold over a three month period of time and the test 
set contains several millions of items sold a few days after the training period. 
More details about the dataset can be found in \cite{shrs12}.   
eBay organizes items into a six-level category structure where there are 39 top level nodes
called \emph{meta categories} and 20K$+$ bottom level nodes called \emph{leaf categories}.
The dataset is generated when users provide 
the titles of items they intend to sell on eBay.
Each title is limited to 50 characters, based on which
the user gets recommendation of some \emph{leaf categories} the item should belong to.
Such categorization of the item helps 
a seller list an item in the correct branch of the product list, thereby allowing a buyer more easily search through a 
list of few million items sold via eBay every single day.
A carefully designed $k$-Nearest Neighbor ($k$-NN) classifier (with the help of improved search engine algorithms) 
categorizes each of the items in less than 100 ms \cite{shrs12}. 
However, due to the large number of categories (20K), items belonging to similar types of 
categories often get misclassified. 

To avoid such confusion, 
larger categories are formed by aggregating examples from categories 
which are relatively difficult to separate. Such aggregation is easy once
the confusion matrix of the classification, obtained from a development dataset, is partitioned and 
strongly connected vertices (each vertex representing one of 20K \emph{leaf categories}) 
are identified from the confusion graph, thereby forming a set of cliques which represent the 
large categories. Note that the large categories so discovered might not at all follow the internal 
hierarchy that is maintained. Next, clustering is performed with examples 
belonging to each of the large categories and the clustering results, along with the predictions from 
$k$-NN classification, are fed to \textbf{BC\textsuperscript{3}E}
(and also to its competitors \emph{i.e.} \textbf{C\textsuperscript{3}E}, \textbf{BGCM}, and \textbf{LWE}).
The idea here is to first reduce the classification space 
and then use unsupervised information to refine the 
predictions from $k$-NN on a smaller number of categories. The number of \emph{leaf categories} belonging to such large categories 
usually varies between 4-10.

However, the dataset is very dynamic and, typically over a span of three months, 20\% of new words 
are added to the existing vocabulary. One can retrain the existing $k$-NN classifier every 
three months, but the training process requires 
collecting new labeled data which is time consuming and expensive. 
One can additionally design classifiers to segregate examples belonging to each of the large 
categories. However, such approach might not improve much upon the performance of the initial 
$k$-NN classifier if the data changes so frequently.  
Therefore, we require a system that can adaptively predict 
newer examples without retraining the existing classifier or employing 
another set of classification algorithms. \textbf{BC\textsuperscript{3}E} 
is useful in such settings. The parameter $\delta$ can adjust the weights of prediction from 
classifiers and unsupervised information. As the results reported in Table \ref{table3} reveal, 
as long as the classification performance is not that poor, \textbf{BC\textsuperscript{3}E} can improve on 
the performance of $k$-NN using the clustering ensemble. 

\scriptsize
\begin{table*}[htbp]
\centering
\begin{tabular}{|l|c|c|c|c|c|c|c|c|}
\hline
\hline
Group ID & $|\mathcal{X}|$ & $k$-NN & BGCM & LWE & C\textsuperscript{3}E-Ideal & BC\textsuperscript{3}E\\ \hline
42 & 1299 & 64.90 & 73.78 ($\pm$ 0.94) & 76.86 ($\pm$ 1.01) & 83.99 ($\pm$ 0.41) & 83.68 ($\pm$ 1.09)\\ \hline
84 & 611 & 63.67 & 69.23 ($\pm$ 0.17) & 75.24 ($\pm$ 0.26) & 81.18 ($\pm$ 0.16) & 76.27 ($\pm$ 1.31)\\ \hline
86 & 2381 & 77.66 & 84.33 ($\pm$ 2.74) & 83.29 ($\pm$ 1.02) & 92.78 ($\pm$ 0.35) & 87.20 ($\pm$ 0.91) \\ \hline
67 & 789 & 72.75 & 72.75 ($\pm$ 0.07) & 78.03 ($\pm$ 0.72) & 82.64 ($\pm$ 0.82) & 81.75 ($\pm$ 1.37) \\ \hline
52 & 1076 & 76.95 & 77.01 ($\pm$ 1.18) & 77.49 ($\pm$ 1.41) & 88.38 ($\pm$ 0.22) & 85.04 ($\pm$ 2.14)\\ \hline
99 & 827 & 84.04 & 85.12 ($\pm$ 0.52) & 86.90 ($\pm$ 0.92) & 91.54 ($\pm$ 0.27) & 91.17 ($\pm$ 0.82) \\ \hline
48 & 3445 & 86.33 & 86.19 ($\pm$ 0.25) & 90.38 ($\pm$ 1.03) & 92.71 ($\pm$ 0.31) & 92.71 ($\pm$ 1.16) \\ \hline
94 & 440 & 79.32 & 81.08 ($\pm$ 0.73) & 82.52 ($\pm$ 0.83) & 85.45 ($\pm$ 0.09) & 85.45 ($\pm$ 0.79) \\ \hline
35 & 4907 & 82.41 & 82.10 ($\pm$ 0.37) & 85.08 ($\pm$ 1.39) & 88.16 ($\pm$ 0.17) & 88.22 ($\pm$ 1.21) \\ \hline
45 & 1952 & 74.80 & 73.12 ($\pm$ 0.81) & 73.64 ($\pm$ 1.68) & 84.32 ($\pm$ 0.23) & 77.97 ($\pm$ 0.47) \\ \hline
\hline
\end{tabular}
\label{table3}
\caption{Performance of \textbf{BC\textsuperscript{3}E} on text classification data --- Avg. Accuracies $\pm$(Standard Deviations).}
\end{table*}
\normalsize

The column ``Group ID'' denotes anonymized groups representing different large categories. $|\mathcal{X}|$ shows the number of examples 
in the test data.
The column ``C\textsuperscript{3}E-Ideal'' shows the performance of \textbf{C\textsuperscript{3}E} if 
the correct tuning parameter for \textbf{C\textsuperscript{3}E} were known. 
For a transfer learning problem, estimating such tuning 
parameter requires some labeled data from the target set which is not available in our setting. If the tuning parameter is 
chosen from cross-validation on the training data, the final prediction on target set can get affected adversely if the underlying 
distribution changes (and in fact it does in our experiments). 
Therefore, we need to adopt a fail-safe approach where we can do 
at least as good as the $k$-NN prediction. The results reveal that \textbf{BC\textsuperscript{3}E} significantly 
outperforms \textbf{BGCM} and \textbf{LWE}, and sometimes achieves as 
good a performance as \textbf{C\textsuperscript{3}E}-Ideal (\emph{i.e.} when correct tuning parameter of \textbf{C\textsuperscript{3}E} is known).
The performance of \textbf{C\textsuperscript{3}E}-Ideal can essentially be considered 
as the best accuracy one could achieve from the given inputs (\emph{i.e.} class and cluster labels) using 
other existing algorithms --- \textbf{BGCM}, \textbf{LWE}, \textbf{C\textsuperscript{3}E} --- that work on the same design 
space. Though \textbf{BGCM} has a tuning parameter, its variation did not affect performance much and we just report 
results corresponding to unity value of this parameter. 

\subsection{Semi-supervised Learning:}

Six datasets are used in our experiments for semi-supervised learning: \textit{Half-Moon} (a synthetic dataset with two half circles 
representing two classes), \textit{Circles} 
(another synthetic dataset that has two-dimensional instances that form 
two concentric circles --- one for each class), and four datasets from the \textit{Library for Support Vector 
Machines} --- \textit{Pima Indians Diabetes}, \textit{Heart}, \textit{German Numer}, and \textit{Wine}.
In order to simulate semi-supervised settings where there is a very limited amount of labeled instances, 
small percentages (see the values reported in Table \ref{table2}) of the instances are randomly selected for training, whereas the remaining instances 
are used for testing (target set). 
We perform 20 trials for every dataset. 
For running experiments with \textbf{BGCM}, and \textbf{C\textsuperscript{3}E}, 
the parameters reported in \cite{galf09} and \cite{achg12a} are used respectively. 
The parameters of \textbf{BC\textsuperscript{3}E} are initialized randomly and approximately 10 EM iterations are 
enough to get the results reported in Table \ref{table2}. 
The classifier ensemble consists of decision tree (C4.5), linear discriminant, and generalized logistic regression.  
Cluster ensembles are generated by means of multiple runs of $k$-means \cite{achg12a}. 
\textbf{LWE} \cite{gafj08} is better suited for transfer learning applications and hence has been left out from 
comparison. The column ``Best'' in Table \ref{table2} refers to the performance of the best 
classifier in the ensemble.
Note that \textbf{BC\textsuperscript{3}E} has superior performance for the most difficult problems, where one has an incentive 
to use a more complex mechanism.
Most importantly, \textbf{BC\textsuperscript{3}E} has the privacy preserving 
property not present in any of its counterparts.


\scriptsize
\begin{table*}[htbp]
\centering
\begin{tabular}{|l|c|c|c|c|c|c|c|c|}
\hline
\hline
Dataset (\% of tr. data) & \multicolumn{1}{l|}{$|\mathcal{X}|$} & Ensemble & Best & BGCM & C\textsuperscript{3}E & BC\textsuperscript{3}E\\ \hline

Half-moon(2\%) & 784 & 92.53$(\pm1.83)$ & 93.02$(\pm0.82)$ & 92.16$(\pm1.47)$ & \textbf{99.64}$(\pm0.08)$ & 98.23$(\pm2.03)$\\ \hline
Circles(2\%) & 1568 & 60.03$(\pm8.44)$ & 95.74$(\pm5.15)$   & 78.67$(\pm0.54)$   & \textbf{99.61}$(\pm0.83)$ & 97.91$(\pm0.74)$ \\ \hline
Pima(2\%)  & 745 & 68.16$(\pm5.05)$ &  69.93$(\pm3.68)$  & 69.21$(\pm4.83)$  &  70.31$(\pm4.44)$ & \textbf{72.83}$(\pm0.49)$\\ \hline   
Heart(7\%)  &  251 & 77.77$(\pm2.55)$   & 79.22$(\pm2.20)$   & 82.78$(\pm4.82)$    & \textbf{82.85}$(\pm5.25)$ & 82.53$(\pm1.14)$\\ \hline
G. Numer(10\%)  &  900 & 70.96$(\pm1.00)$ & 70.19$(\pm1.52)$ &    73.70$(\pm1.06)$  &  74.44$(\pm3.44)$ & \textbf{74.61}$(\pm1.62)$\\ \hline
Wine(10\%)  &  900 & 79.87$(\pm5.68)$ & 80.37$(\pm5.47)$ &  75.37$(\pm13.66)$  & \textbf{83.62}$(\pm6.27)$ & 82.20$(\pm1.07)$\\ \hline
\hline
\end{tabular}
\label{table2}
\caption{Comparison of \textbf{BC\textsuperscript{3}E} with \textbf{C\textsuperscript{3}E} and \textbf{BGCM} --- Avg. Accuracies $\pm$(Standard Deviations).}
\end{table*}
\normalsize

\section{Conclusion and Future Work}
\label{conclusion}

The \textbf{BC\textsuperscript{3}E} model proposed in this paper has been shown to be useful for 
difficult non-transductive semisupervised and transfer learning problems.
A good trade-off between accuracy and 
privacy has also been established empirically -- a property absent in 
any of \textbf{BC\textsuperscript{3}E}'s competitors.
With minor modification, \textbf{BC\textsuperscript{3}E} can also handle 
soft outputs from classification and clustering ensembles which can further improve the results.

\bibliographystyle{siam}

\end{document}


\title{\Large Supplement to Probabilistic Combination of Classifier and Cluster Ensembles for Non-transductive Learning}
\author{Ayan Acharya\thanks{University of Texas at Austin, Austin, TX, USA. Email: \{aacharya@, ghosh@ece\}.utexas.edu}\\
\and
Eduardo R. Hruschka\thanks{University of Sao Paulo at Sao Carlos, Brazil. Email: erh@icmc.usp.br}
\and
Joydeep Ghosh$^{\ast}$
\and
Badrul Sarwar\thanks{eBay Research Lab, San Jose, CA, USA. Email: \{bsarwar, jruvini\}@ebay.com}
\and
Jean-David Ruvini$^{\ddagger}$
}
\date{}

\maketitle

\begin{table*}[htbp]
\centering
\begin{tabular}{|c|c||c|c|}
\hline
\hline
 Edge No. & Message & Edge No. & Message\\ \hline
 1 & $\vect{\theta}^m\backslash \{\beta_{mij}\}$ & 2 & partial sums\\ \hline 
 3 & $\vect{\theta}^m\backslash \{\beta_{mij}\}$ & 4 & partial sums\\ \hline
 5 & $\vect{\theta}^m\backslash \{\beta_{mij}\}$ & 6 & partial sums\\ \hline 
 7 & $\vect{\theta}^m\backslash \{\beta_{mij}\}$ & 8 & partial sums\\ \hline
 9 & $\left\{\displaystyle\sum_{m\in G_{1}}\displaystyle\sum_{i=1}^{k}\phi_{nmi}\epsilon_{ni},
 \displaystyle\sum_{l\in G_{1}}\displaystyle\sum_{i=1}^{k}w_{1nli}\mu_{ni}\right\}_{n\in\mathcal{X}_{1}}$  & 10 & 
 $\{\vect{\epsilon}_{n}\}_{n\in\mathcal{X}_{1}}$\\ \hline
 11 & $\left\{\displaystyle\sum_{m\in G_{2}\cup G_{3}}\displaystyle\sum_{i=1}^{k}\phi_{nmi}\epsilon_{ni},
 \displaystyle\sum_{l\in G_{2}\cup G_{3}}\displaystyle\sum_{i=1}^{k}w_{1nli}\mu_{ni}\right\}_{n\in\mathcal{X}_{1}}$ & 12 & 
  $\{\vect{\epsilon}_{n}\}_{n\in\mathcal{X}_{1}}$\\ \hline
 13 & $\left\{\displaystyle\sum_{m\in G_{2}\cup G_{3}}\displaystyle\sum_{i=1}^{k}\phi_{nmi}\epsilon_{ni},
 \displaystyle\sum_{l\in G_{2}\cup G_{3}}\displaystyle\sum_{i=1}^{k}w_{1nli}\mu_{ni}\right\}_{n\in\mathcal{X}_{2}}$ & 14 & 
  $\{\vect{\epsilon}_{n}\}_{n\in\mathcal{X}_{2}}$ \\ \hline
 15 & $\left\{\displaystyle\sum_{m\in G_{1}}\displaystyle\sum_{i=1}^{k}\phi_{nmi}\epsilon_{ni},
 \displaystyle\sum_{l\in G_{1}}\displaystyle\sum_{i=1}^{k}w_{1nli}\mu_{ni}\right\}_{n\in\mathcal{X}_{2}}$ & 16 & 
  $\{\vect{\epsilon}_{n}\}_{n\in\mathcal{X}_{2}}$\\ \hline
 17 & $\left\{\displaystyle\sum_{m\in G_{1}}\displaystyle\sum_{i=1}^{k}\phi_{nmi}\epsilon_{ni},
 \displaystyle\sum_{l\in G_{1}}\displaystyle\sum_{i=1}^{k}w_{1nli}\mu_{ni}\right\}_{n\in\mathcal{X}_{3}}$ & 18 & 
 $\{\vect{\epsilon}_{n}\}_{n\in\mathcal{X}_{3}}$ \\ \hline
 19 & $\left\{\displaystyle\sum_{m\in G_{2}\cup G_{3}}\displaystyle\sum_{i=1}^{k}\phi_{nmi}\epsilon_{ni},
 \displaystyle\sum_{l\in G_{2}\cup G_{3}}\displaystyle\sum_{i=1}^{k}w_{1nli}\mu_{ni}\right\}_{n\in\mathcal{X}_{3}}$ & 20 & $\{\vect{\epsilon}_{n}\}_{n\in\mathcal{X}_{3}}$\\ \hline
 21 & $\left\{\displaystyle\sum_{m\in G_{1}\cup G_{2}}\displaystyle\sum_{i=1}^{k}\phi_{nmi}\epsilon_{ni},
 \displaystyle\sum_{l\in G_{1}\cup G_{2}}\displaystyle\sum_{i=1}^{k}w_{1nli}\mu_{ni}\right\}_{n\in\mathcal{X}_{4}}$ & 22 & $\{\vect{\epsilon}_{n}\}_{n\in\mathcal{X}_{4}}$\\ \hline
 23 & $\left\{\displaystyle\sum_{m\in G_{3}}\displaystyle\sum_{i=1}^{k}\phi_{nmi}\epsilon_{ni},
 \displaystyle\sum_{l\in G_{3}}\displaystyle\sum_{i=1}^{k}w_{1nli}\mu_{ni}\right\}_{n\in\mathcal{X}_{4}}$ & 24 & 
 $\{\vect{\epsilon}_{n}\}_{n\in\mathcal{X}_{4}}$ \\ \hline
 25 & $\left\{\displaystyle\sum_{n\in\mathcal{X}_{1}}\phi_{nmi}w_{2nmj}\right\}_{m\in G_{1}}$ &
 26 & $\{\beta_{mij}:m\in G_{1}\}$\\ \hline
 27 & $\left\{\displaystyle\sum_{n\in \mathcal{X}_{2}\cup\mathcal{X}_{3}}\phi_{nmi}w_{2nmj}\right\}_{m\in G_{1}}$ &
 28 & $\{\beta_{mij}:m\in G_{1}\}$\\ \hline
 29 & $\left\{\displaystyle\sum_{n\in\mathcal{X}_{4}}\phi_{nmi}w_{2nmj}\right\}_{m\in G_{1}}$ &
 30 & $\{\beta_{mij}:m\in G_{1}\}$\\ \hline
 31 & $\left\{\displaystyle\sum_{n\in \mathcal{X}_{1}\cup\mathcal{X}_{2}}\phi_{nmi}w_{2nmj}\right\}_{m\in G_{2}}$ &
 32 & $\{\beta_{mij}:m\in G_{2}\}$\\ \hline
 33 & $\left\{\displaystyle\sum_{n\in\mathcal{X}_{3}}\phi_{nmi}w_{2nmj}\right\}_{m\in G_{2}}$ &
 34 & $\{\beta_{mij}:m\in G_{2}\}$\\ \hline
 35 & $\left\{\displaystyle\sum_{n\in \mathcal{X}_{4}}\phi_{nmi}w_{2nmj}\right\}_{m\in G_{2}}$ &
 36 & $\{\beta_{mij}:m\in G_{2}\}$\\ \hline
 37 & $\left\{\displaystyle\sum_{n\in\mathcal{X}_{1}\cup\mathcal{X}_{2}}\phi_{nmi}w_{2nmj}\right\}_{m\in G_{3}}$ &
 38 & $\{\beta_{mij}:m\in G_{3}\}$\\ \hline
 39 & $\left\{\displaystyle\sum_{n\in \mathcal{X}_{3}}\phi_{nmi}w_{2nmj}\right\}_{m\in G_{3}}$ &
 40 & $\{\beta_{mij}:m\in G_{3}\}$\\ \hline
 41 & $\left\{\displaystyle\sum_{n\in\mathcal{X}_{4}}\phi_{nmi}w_{2nmj}\right\}_{m\in G_{3}}$ &
 42 & $\{\beta_{mij}:m\in G_{3}\}$\\ \hline

\hline
\hline
\end{tabular}
\caption{Edges and Messages Carried in Arbitrarily Partitioned Ensemble}
\label{tablearbd}
\end{table*}

%% file: BC3E_fig.tex
\begin{tikzpicture}[ >=stealth]
	
           \draw [thick] (0.00,0.00) rectangle (4.00,5.00);
	   \draw (3.80, 0.20) node {$N$};
  
           \draw [thick] (0.25,0.50) rectangle (1.75,3.50);
	   \draw (1.6, 0.7) node {$r_{1}$};
             
           \draw [thick] (2.25,0.50) rectangle (3.75,3.50);
	   \draw (3.6, 0.7) node {$r_{2}$};

           \draw [thick] (1,4.25) circle (0.4cm); 

           \draw [thick] (3,4.25) circle (0.4cm); 
           \draw [thick] (3,2.75) circle (0.4cm); 
           \filldraw [gray] (3,1.25) circle (0.4cm); 
           \filldraw [gray] (1,1.25) circle (0.4cm); 
           \draw (3,4.25) node {$\boldsymbol{\theta}$};
           \draw (1,4.25) node {$\vect{y}$};
	   \draw (3,2.75) node {$\vect{z}$};
	   \draw (3,1.25) node {$\vect{w}_{2}$};
	   \draw (1,1.25) node {$\vect{w}_{1}$};
           \draw (5.25,1.25) node {$\boldsymbol{\beta}$}; 

           \filldraw [black] (5,1.25) circle (0.1cm); 
           \filldraw [black] (3,5.50) circle (0.1cm); 
           \filldraw [black] (1,5.50) circle (0.1cm); 
           
           \draw (3, 5.80) node {$\boldsymbol{\delta^{2}}$};
           \draw (1, 5.80) node {$\boldsymbol{\mu, \sigma^{2}}$};

           \draw[->, thick] (3,3.85) -- (3,3.15);
           \draw[->, thick] (3,2.35) -- (3,1.65);
           \draw[->, thick] (1,3.85) -- (1,1.65);

           \draw[->, thick] (1.4,4.25) -- (2.6,4.25);
           \draw[->, thick] (4.9,1.25) -- (3.4,1.25);
           \draw[->, thick] (3,5.40) -- (3,4.65);
           \draw[->, thick] (1,5.40) -- (1,4.65);

           \draw [thick] (4.25,0.50) rectangle (5.75,2.00);
	   \draw (5.20, 0.70) node {$r_{2}\times k$};

\end{tikzpicture}

%% file: algorithm_BC3E.tex
\begin{algorithm}[ht]
\caption{Learning \textbf{BC\textsuperscript{3}E}}
\label{algo:LearnBC3E}
\begin{footnotesize}
\begin{algorithmic}
\STATE  {\bf Input:} $\vect{W}$.
\STATE  {\bf Output:} $\vect{\theta}^{m}, \{\vect{\mu}_{n}\}_{n=1}^{N}$.
\STATE
\STATE Initialize $\vect{\theta}^{m}$, $\{\vect{\zeta}_{n}\}_{n=1}^{N}$.
\STATE Until Convergence
\STATE {\bf E-Step}
\STATE Until Convergence
\STATE 1.  Update $\kappa_{n}$ using Eq.~\eqref{updatekappa} $\forall  n\in\{1,2,\cdots,N\}$.
\STATE 2.  Update $\xi_{n}$ using Eq.~\eqref{updatexi} $\forall  n\in\{1,2,\cdots,N\}$.
\STATE 3.  Update $\phi_{nmi}$ using Eq.~\eqref{updatephi} $\forall n,m,i$. Normalize $\vect{\phi}_{nm}$.
\STATE 4.  Maximize \eqref{updatemun} w.r.t. $\vect{\mu}_{n}$ $\forall n$.
\STATE 5.  Maximize \eqref{updatesigman2} w.r.t. $\vect{\sigma}_{n}^2$ $\forall n$ s.t. $\vect{\sigma}_{n}^2 \ge \vect{0}$.
\STATE 6. Maximize \eqref{updateepsilon} w.r.t. $\vect{\epsilon}_{n}$ $\forall n$. 
\STATE 7. Maximize \eqref{updatedeltan2} w.r.t. $\vect{\delta}_{n}^{2}$ $\forall n$ s.t. $\vect{\delta}_{n}^2 \ge 0$.
\STATE {\bf M-Step}
\STATE 8. Update $\vect{\mu}$ using Eq.~\eqref{updatemu}.
\STATE 9. Update $\delta^{2}$ using Eq.~\eqref{updatedelta2}.
\STATE 10. Update $\beta_{mij}$ using Eq.~\eqref{updatebeta} $\forall m,i,j$. Normalize $\vect{\theta}_{mi}$.
\STATE 11. Maximize \eqref{updatesigma2} w.r.t. $\vect{\sigma}^2$ s.t. $\vect{\sigma}^2 \ge \vect{0}$. 
\end{algorithmic}
\end{footnotesize}
\end{algorithm}

%% file: figarbd.tex
\begin{tikzpicture}[ >=stealth]

\tikzstyle{redfill} = [fill=red,fill opacity=0.7]
\tikzstyle{bluefill} = [fill=blue,fill opacity=0.7]
\tikzstyle{greenfill} = [fill=green,fill opacity=0.7]
\tikzstyle{violetfill} = [fill=violet,fill opacity=0.7]

\filldraw [redfill] (0,0) rectangle (3,2);
\filldraw [bluefill] (0,2) rectangle (2,5);
\filldraw [greenfill] (3,0) rectangle (5,2);
\filldraw [violetfill] (2,2) rectangle (5,4);
\filldraw [gray] (2,4) rectangle (5,7);

\draw [thick] (0.00,0.00) rectangle (5.00,7.00);
\draw [thick] (0.00,0.00) rectangle (5.00,6.00);
\draw [thick] (0.00,0.00) rectangle (5.00,5.00);
\draw [thick] (0.00,0.00) rectangle (5.00,4.00);
\draw [thick] (0.00,0.00) rectangle (5.00,3.00);
\draw [thick] (0.00,0.00) rectangle (5.00,2.00);
\draw [thick] (0.00,0.00) rectangle (5.00,1.00);

\draw [thick] (0.00,0.00) rectangle (1.00,7.00);
\draw [thick] (0.00,0.00) rectangle (2.00,7.00);
\draw [thick] (0.00,0.00) rectangle (3.00,7.00);
\draw [thick] (0.00,0.00) rectangle (4.00,7.00);

\draw [decorate,decoration={brace,amplitude=5pt}]
   (-0.5,0)  -- (-0.5,2) 
   node [black,midway,xshift=-12pt] {\footnotesize $\mathcal{X}_{4}$};

\draw [decorate,decoration={brace,amplitude=5pt}]
   (-0.5,2)  -- (-0.5,4) 
   node [black,midway,xshift=-12pt] {\footnotesize $\mathcal{X}_{3}$};

\draw [decorate,decoration={brace,amplitude=5pt}]
   (-0.5,4)  -- (-0.5,5) 
   node [black,midway,xshift=-12pt] {\footnotesize $\mathcal{X}_{2}$};

\draw [decorate,decoration={brace,amplitude=5pt}]
   (-0.5,5)  -- (-0.5,7) 
   node [black,midway,xshift=-12pt] {\footnotesize $\mathcal{X}_{1}$};

\draw [decorate,decoration={brace,amplitude=5pt}]
   (0,7.5)  -- (2,7.5) 
   node [black,midway,yshift=12pt] {\footnotesize $G_{1}$};

\draw [decorate,decoration={brace,amplitude=5pt}]
   (2,7.5)  -- (3,7.5) 
   node [black,midway,yshift=12pt] {\footnotesize $G_{2}$};

\draw [decorate,decoration={brace,amplitude=5pt}]
   (3,7.5)  -- (5,7.5) 
   node [black,midway,yshift=12pt] {\footnotesize $G_{3}$};

\end{tikzpicture}

%% file: case4fig.tex
\begin{tikzpicture}[ >=stealth]
\label{case4Fig}

\tikzstyle{redfill} = [fill=red,fill opacity=0.5];
\tikzstyle{bluefill} = [fill=blue,fill opacity=0.5];
\tikzstyle{greenfill} = [fill=green,fill opacity=0.5];
\tikzstyle{violetfill} = [fill=violet,fill opacity=0.5];           
           
\node [circle, draw] (a) {S};
\node [right=0mm of a] {\ref{updatemu}, \ref{updatedelta2}, \ref{updatesigma2}};

\node [circle, draw] (c) [below left=2cm of a] {AS$_2$};
\node [circle, draw] (d) [below right=2cm of a] {AS$_3$};
\node [circle, draw] (b) [left=2cm of c] {AS$_1$};
\node [circle, draw] (e) [right=2cm of d] {AS$_4$};

\node [circle, draw] (h) [below=5cm of a,xshift=-1.5cm, fill=blue] {C$_3$};
\node [circle, draw, fill=gray] (f) [left=1.5cm of h] {C$_2$};
\node [circle, draw] (g) [left=1.5cm of f] {C$_1$};

\node [circle, draw, fill=violet] (i) [below=5cm of a,xshift=1.5cm] {C$_4$};
\node [circle, draw, fill=red] (j) [right=1.5cm of i] {C$_5$};
\node [circle, draw, fill=green] (k) [right=1.5cm of j] {C$_6$};

\node [circle, draw] (l) [below=8cm of a] {AC$_2$};
\node [circle, draw] (m) [left=3cm of l] {AC$_1$};
\node [circle, draw] (n) [right=3cm of l] {AC$_3$};

\draw [->] (a) to [bend left=0] node [auto, swap, very near start] {$1$} (b);
\draw [->] (b) to [bend left=0] node [auto, very near start] {$2$} (a);
\draw [->] (a) to [bend left=0] node [auto, very near start] {$3$} (c);
\draw [->] (c) to [bend left=0] node [auto, very near start] {$4$} (a);
\draw [->] (a) to [bend left=0] node [auto, swap, very near start] {$5$} (d);
\draw [->] (d) to [bend left=0] node [auto, very near start] {$6$} (a);
\draw [->] (a) to [bend left=0] node [auto, very near start, yshift=-0.2cm] {$7$} (e);
\draw [->] (e) to [bend left=0] node [auto, very near start] {$8$} (a);

\draw [->] (g) to [bend left=0] node [auto, very near start] {$9$} (b);
\draw [->] (b) to [bend left=0] node [auto, swap, very near start] {$10$} (g);
\draw [->] (f) to [bend left=0] node [auto, very near start] {$11$} (b);
\draw [->] (b) to [bend left=0] node [auto, very near start] {$12$} (f);
\draw [->] (f) to [bend left=0] node [auto, very near start] {$13$} (c);
\draw [->] (c) to [bend left=0] node [auto, swap, very near start] {$14$} (f);
\draw [->] (h) to [bend left=0] node [auto, very near start] {$15$} (c);
\draw [->] (c) to [bend left=0] node [auto, very near start] {$16$} (h);
\draw [->] (h) to [bend left=0] node [auto, very near start] {$17$} (d);
\draw [->] (d) to [bend left=0] node [auto, swap, very near start] {$18$} (h);
\draw [->] (i) to [bend left=0] node [auto, very near start] {$19$} (d);
\draw [->] (d) to [bend left=0] node [auto, very near start] {$20$} (i);
\draw [->] (j) to [bend left=0] node [auto, very near start] {$21$} (e);
\draw [->] (e) to [bend left=0] node [auto, swap, very near start] {$22$} (j);
\draw [->] (k) to [bend left=0] node [auto, very near start] {$23$} (e);
\draw [->] (e) to [bend left=0] node [auto, very near start] {$24$} (k);

\draw [->] (g) to [bend left=0] node [auto,very near start] {$25$} (m);
\draw [->] (m) to [bend left=0] node [auto,very near start] {$26$} (g);
\draw [->] (h) to [bend left=0] node [auto,very near start] {$27$} (m);
\draw [->] (m) to [bend left=0] node [auto, very near start] {$28$} (h);
\draw [->] (j) to [bend left=0] node [auto, swap, very near start, xshift=0.5cm] {$29$} (m);
\draw [->] (m) to [bend left=0] node [auto, swap, very near start] {$30$} (j);

\draw [->] (f) to [bend left=0] node [auto, swap, very near start] {$31$} (l);
\draw [->] (l) to [bend left=0] node [auto, very near start] {$32$} (f);
\draw [->] (i) to [bend left=0] node [auto, swap, very near start] {$33$} (l);
\draw [->] (l) to [bend left=0] node [auto, very near start] {$34$} (i);
\draw [->] (j) to [bend left=0] node [auto, very near start] {$35$} (l);
\draw [->] (l) to [bend left=0] node [auto, swap, very near start] {$36$} (j);

\draw [->] (f) to [bend left=0] node [auto, very near start, xshift=-0.5cm] {$37$} (n);
\draw [->] (n) to [bend left=0] node [auto, very near start] {$38$} (f);
\draw [->] (i) to [bend left=0] node [auto, swap, very near start] {$39$} (n);
\draw [->] (n) to [bend left=0] node [auto, swap, very near start] {$40$} (i);
\draw [->] (k) to [bend left=0] node [auto, very near start] {$41$} (n);
\draw [->] (n) to [bend left=0] node [auto, very near start] {$42$} (k);

\node [right=0mm of n] {\ref{updatebeta}};
\node [right=0mm of j] {\ref{updatephi}};
\node [right=0mm of c] {\ref{updatekappa}, \ref{updatexi},  \ref{updatedeltan2}, \ref{updatemun}, \ref{updatesigman2}, \ref{updateepsilon}};


%
\end{tikzpicture}